\documentclass[sigconf, nonacm]{acmart}

\newcommand\vldbdoi{XX.XX/XXX.XX}
\newcommand\vldbpages{XXX-XXX}
\newcommand\vldbvolume{18}
\newcommand\vldbissue{10}
\newcommand\vldbyear{2025}
\newcommand\vldbauthors{\authors}
\newcommand\vldbtitle{\shorttitle} 
\newcommand\vldbavailabilityurl{https://github.com/ruc-datalab/AutoPrep}
\newcommand\vldbpagestyle{empty} 

\usepackage{framed}
\usepackage{xspace}
\usepackage{listings}
\usepackage{framed}
\usepackage{multirow}
\usepackage[normalem]{ulem}
\usepackage{algorithm}
\usepackage{algpseudocode}
\usepackage{graphicx}
\usepackage{soul}
\usepackage{textcomp}
\usepackage{xcolor}
\usepackage{multirow}
\usepackage{booktabs}
\usepackage{balance}
\usepackage{tabularx}
\usepackage{subcaption}
\usepackage{bm}
\usepackage{url}
\usepackage{makecell}
\usepackage{wasysym}
\usepackage{marvosym}
\usepackage{enumitem}
\usepackage{pifont}

\newtheorem{example}{Example}

\definecolor{cadmiumgreen}{rgb}{0.0, 0.42, 0.24}
\definecolor{dropred}{rgb}{0.75, 0.22, 0.17}
\newcommand{\eat}[1]{}

\newcounter{mycountermodification}

\newcounter{mycounterexp}
\renewcommand{\themycounterexp}{\arabic{mycounterexp}}
\newcommand{\expnum}{%
    \refstepcounter{mycounterexp}
    \themycounterexp
}

\newcounter{mycounterfinding}

\colorlet{shadecolor}{gray!20}

\definecolor{shadecolor}{RGB}{220,220,220}

\newcommand{\sstab}{\rule{0pt}{8pt}\\[-2.2ex]}

\newcommand{\bi}{\begin{itemize}}
\newcommand{\ei}{\end{itemize}}

\newcommand{\be}{\begin{enumerate}}
\newcommand{\ee}{\end{enumerate}}

\newcommand{\stitle}[1]{\sstab\noindent{\bf #1}}
\newcommand{\etitle}[1]{\vspace{0.5mm}\noindent{\underline{\em #1}}}
\newcommand{\ie}{{\em i.e.,}\xspace}
\newcommand{\eg}{{\em e.g.,}\xspace}

\newcommand{\term}[1]{{\tt #1}}

\newcommand{\sys}{\textsc{AutoPrep}\xspace}

\begin{document}
\pagestyle{\vldbpagestyle}
\title{\sys: Natural Language Question-Aware Data Preparation with a Multi-Agent Framework}

\newcommand{\revise}[1]{{\color{black} #1}}


\author{Meihao Fan}
\affiliation{%
  \institution{Renmin University of China}
}
\email{fmh1art@ruc.edu.cn}

\author{Ju Fan}
\authornote{Ju Fan is the corresponding author.}
\affiliation{%
  \institution{Renmin University of China}
}
\email{fanj@ruc.edu.cn}

\author{Nan Tang}
\affiliation{%
  \institution{HKUST (GZ)}
}
\email{nantang@hkust-gz.edu.cn}

\author{Lei Cao}
\affiliation{%
  \institution{University of Arizona}
}
\email{caolei@arizona.edu}

\author{Guoliang Li}
\affiliation{%
  \institution{Tsinghua University}
}
\email{liguoliang@tsinghua.edu.cn}

\author{Xiaoyong Du}
\affiliation{%
  \institution{Renmin University of China}
}
\email{duyong@ruc.edu.cn}

\begin{abstract}
Answering natural language (NL) questions about tables, known as Tabular Question Answering (TQA), is crucial because it allows users to quickly and efficiently extract meaningful insights from structured data, effectively bridging the gap between human language and machine-readable formats. 
Many of these tables are derived from web sources or real-world scenarios, which require meticulous data preparation (or data prep) to ensure accurate responses. 
However, preparing such tables for NL questions introduces new requirements that extend beyond traditional data preparation. This question-aware data preparation involves specific tasks such as column derivation and filtering tailored to particular questions, as well as question-aware value normalization or conversion, highlighting the need for a more nuanced approach in this context.
%
%
Because each of the above tasks is unique, a single model (or agent) may not perform effectively across all scenarios. In this paper, we propose \textbf{\sys}, a large language model (LLM)-based multi-agent framework that leverages the strengths of multiple agents, each specialized in a certain type of data prep, ensuring more accurate and contextually relevant responses.
Given an NL question over a table, \sys performs data prep through three key components.
{\bf Planner}: Determines a logical plan, outlining a sequence of high-level operations.
{\bf Programmer}: Translates this logical plan into a physical plan by generating the corresponding low-level code.
{\bf Executor}: Executes the generated code to process the table.
%
To support this multi-agent framework, we design a novel Chain-of-Clauses reasoning mechanism for high-level operation suggestion, and a tool-augmented method for low-level code generation.
Extensive experiments on real-world TQA datasets demonstrate that \sys can significantly improve the state-of-the-art TQA solutions through question-aware data preparation.

\end{abstract}
\maketitle
\begingroup\small\noindent\raggedright\textbf{PVLDB Reference Format:}\\
\vldbauthors. \vldbtitle. PVLDB, \vldbvolume(\vldbissue): \vldbpages, \vldbyear.\\
\href{https://doi.org/\vldbdoi}{doi:\vldbdoi}
\endgroup
\begingroup
\renewcommand\thefootnote{}\footnote{\noindent
This work is licensed under the Creative Commons BY-NC-ND 4.0 International License. Visit \url{https://creativecommons.org/licenses/by-nc-nd/4.0/} to view a copy of this license. For any use beyond those covered by this license, obtain permission by emailing \href{mailto:info@vldb.org}{info@vldb.org}. Copyright is held by the owner/author(s). Publication rights licensed to the VLDB Endowment. \\
\raggedright Proceedings of the VLDB Endowment, Vol. \vldbvolume, No. \vldbissue\ %
ISSN 2150-8097. \\
\href{https://doi.org/\vldbdoi}{doi:\vldbdoi} \\
}\addtocounter{footnote}{-1}\endgroup

\ifdefempty{\vldbavailabilityurl}{}{
\vspace{.3cm}
\begingroup\small\noindent\raggedright\textbf{PVLDB Artifact Availability:}\\
The source code, data, and/or other artifacts have been made available at \url{\vldbavailabilityurl}.
\endgroup
}

\section{Introduction}
\label{sec:intro}

Tabular Question Answering (TQA) refers to the task of answering natural language (NL) questions based on provided tables~\cite{DBLP:journals/fcsc/LuZFFCD25, cheng2022hitab, cheng2023binding, pal2023multitabqa}. TQA empowers non-technical users such as domain scientists to easily analyze tabular data and has a wide range of applications, including table-based fact verification~\cite{chen2019tabfact, DBLP:conf/emnlp/GuF0NZ022, DBLP:conf/sigmod/GuFZ0F022} and table-based question answering~\cite{pasupat2015compositional, nan2022fetaqa}.
As TQA requires NL understanding and reasoning over tables, state-of-the-art solutions~\cite{wei2022chain,ye2023large,DBLP:conf/iclr/YaoZYDSN023,cheng2023binding,zhang2024reactable, wang2024chain} mainly rely on large language models (LLMs).  

\begin{figure}[t]
    \centering
    \includegraphics[width=0.99\columnwidth]{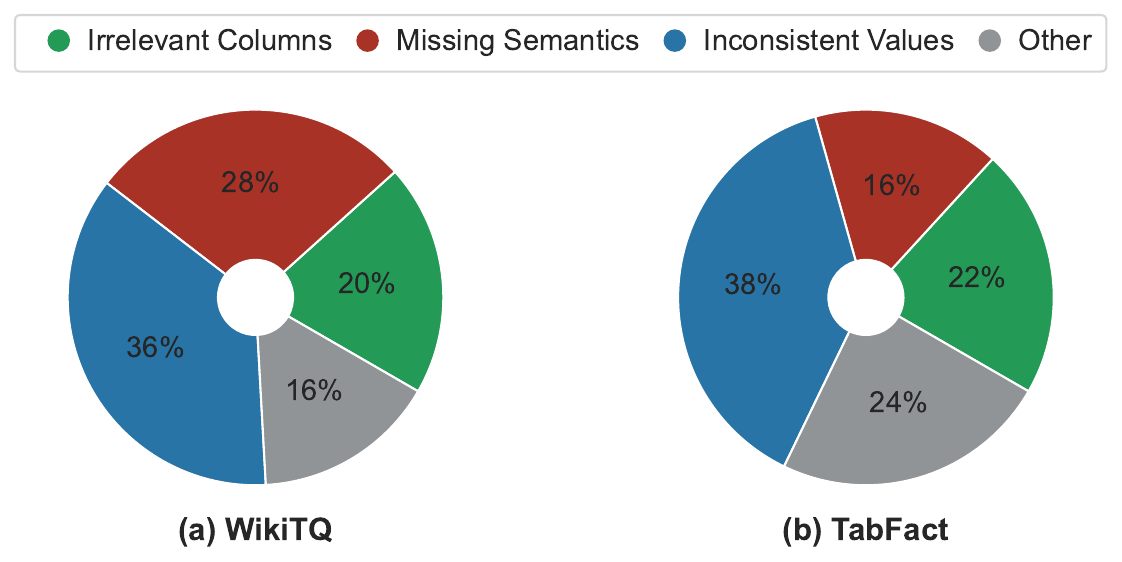}
    \vspace{-1em} 
    \caption{\revise{An error analysis of LLM-based TQA (using GPT-4) on two well-adopted datasets.} 
}
    \label{fig:error_instance_record}
    \vspace{-1em}
\end{figure}

As many tables in TQA originate from web sources or real-world data, they demand meticulous \textbf{data preparation} (or data prep) to produce accurate answers.
Figure~\ref{fig:error_instance_record} shows an error analysis of an LLM-based approach (using GPT-4) across two TQA tasks: table-based question answering on the WikiTQ dataset~\cite{pasupat2015compositional} and table-based fact verification on the TabFact dataset~\cite{chen2019tabfact} \revise{(More details of the error analysis can be found in our technical report~\cite{technical_report})}.
\revise{
	The results indicate that 84\% and 76\% of the errors stem from inadequately addressing data prep issues, including \emph{missing semantics}, \emph{inconsistent values}, and \emph{irrelevant columns}, as illustrated as follows.
}


\begin{figure*}[t]
    \centering
    
    \begin{subfigure}{\textwidth}
        \centering
        \includegraphics[width=0.99\textwidth]{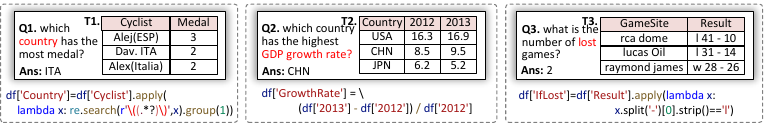}
        \captionsetup{skip=0.2em}
        \caption{\revise{Missing Semantics: Required semantic information for NL questions is not explicitly present in the tables.}}
        \vspace{0.5em}
        \label{fig:issue_instances_incomplete}
    \end{subfigure}
    
    \begin{subfigure}{\textwidth}
        \centering
        \includegraphics[width=0.99\textwidth]{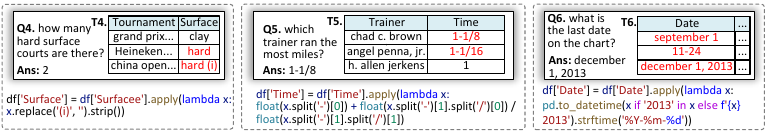}
        \captionsetup{skip=0.2em}
        \caption{\revise{Value Inconsistency: Heterogeneous data representations or formats across table values, where normalization is question-dependent.}}
        \vspace{0.5em}
        \label{fig:issue_instances_inconsistency}
    \end{subfigure}
    
    \begin{subfigure}{\textwidth}
        \centering
        \includegraphics[width=0.99\textwidth]{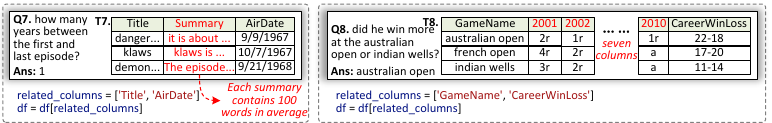}
        \captionsetup{skip=0.2em}
        \caption{\revise{Irrelevant Columns: Presence of unnecessary columns that increase complexity and induce errors for LLMs.
        	}}
        \vspace{0.5em}
        \label{fig:issue_instances_redundancy}
    \end{subfigure}
    
    \vspace{-1em}
    \caption{\revise{Examples of Data Preparation Issues for Table Question Answering.}}
    \label{fig:issue_instances}
    \vspace{-1em}
\end{figure*}

\revise{
\etitle{\textbf{(1) Missing Semantics.}}
This data prep issue arises when a table lacks the necessary \emph{semantics} to address the specific requirements of the NL question. That is, although some columns in the table may be related to the question, they do not directly provide the required semantic information. As shown in Figure~\ref{fig:issue_instances_incomplete}, the semantics needed for the NL questions, such as \emph{country} and \emph{GDP}, are not explicitly present in the tables. Therefore, to ensure accurate responses in TQA, it is essential to perform \textbf{column derivation} from existing columns, such as extracting country information from text or applying mathematical operations across multiple columns.
}

\revise{
\etitle{\textbf{(2) Inconsistent Values.}}  
This data preparation issue arises when the values within a particular column are inconsistent across different records due to varying data representations (\eg inconsistent date formats), as illustrated in Figure~\ref{fig:issue_instances_inconsistency}.  
This situation necessitates \textbf{column normalization}, which must be question-aware, as different questions may require different types of normalization.  
For instance, as $Q_5$ requires comparing numbers in $T_5$ but the corresponding columns are of string type, it is necessary to convert the strings in fraction format into numerical values.
}

\revise{
\etitle{\textbf{(3) Irrelevant Columns.}}
In some tables, although many columns are available, only a few are actually relevant to the NL question.
For example, as shown in Figure~\ref{fig:issue_instances_redundancy}, $Q_7$ does not require the \term{Summary} column in table $T_7$. Similarly, in $T_8$, although there are 12 columns, only 2 are needed to answer $Q_8$.
Thus, \textbf{column filtering}, \ie retaining only the columns directly related to the question, is crucial for TQA, as irrelevant columns add complexity and can mislead LLMs into producing incorrect answers.
%
}

\revise{
	This highlights the critical role of thorough data prep in ensuring accurate TQA.
	Note that, although other issues like missing values and duplicates are common in real-world data, they are less prevalent than the three highlighted issues in current TQA benchmarks.
	Nevertheless, we recognize their significance and leave the exploration of broader data prep challenges to future work.
}

\stitle{Question-Aware Data Preparation.}
In this paper, we thus propose to study a new problem, namely \emph{question-aware data preparation}.
\revise{
The novelty of this problem lies in the need to {deeply understand the semantics of the NL question in order to guide data prep operations over the structured table}. Unlike traditional data prep, which is performed \emph{offline} and independent of downstream tasks, question-aware data prep is conducted \emph{online} and must dynamically tailor the table to the specific demands of the question. This introduces new challenges due to the diversity and ambiguity of NL questions and the heterogeneity of tabular data, requiring precise semantic alignment between the question and the structured table.
%
For example, as shown in Figure~\ref{fig:issue_instances_incomplete}, given question $Q_1$ that requests country-specific information from a text column, table $T_1$ needs to be transformed to extract and create a new \term{Country} column, which may not be considered in traditional data prep.
}


%


%
However, building a system to support question-aware data prep is challenging due to \revise{the semantic alignment between the table and the NL question}. Specifically, different questions may demand different types of data prep, \eg derivation, normalization, or filtering. Furthermore, even if facing the same type of data prep issue, different questions may require different ways to handle it. For example, the method that normalizes the date to a unified format is clearly different from that of normalizing strings to integers.
Although LLMs show promise for interpreting semantics in NL questions, simply applying them to address all potential data prep issues has shown ineffective, often leading to false negatives and false positives, as illustrated in Section~\ref{subsec:naive_method_for_question_aware_data_prep}.



\stitle{\sys: A Hierarchical, Multi-Stage Approach.} 
To address the challenges, we propose \textbf{\sys}, which features two key ideas. First, drawing inspiration from modern DBMS, particularly the distinction between logical operations and their physical implementations, \sys separates high-level, logical data prep operations from the concrete methods used for execution. Specifically, it introduces a planning stage that generates a {\it logical plan} for each question, consisting of a sequence of high-level data prep operations tailored to the question's needs, such as column derivation, normalization and filtering shown in Figure~\ref{fig:issue_instances}. In the next stage, \sys maps these logical operations to the corresponding physical implementations. This separation allows \sys to break down complex data prep tasks into smaller, more manageable sub-tasks, making the process easier to solve. Moreover, this modular design makes \sys extensible, enabling the development of specialized implementations for each type of data prep operation or the introduction of new operation types.
%

Second, unlike conventional DBMSs, \sys does not predefine physical operations for each logical operation. Instead, for each data prep operation in the logical plan, \sys generates a physical implementation that is specialized to the specific question {\it on the fly}. By considering the unique requirements of the NL question, this customized implementation ensures that the data preparation is closely aligned with the needs of different questions.


Building on the above insights, we design the \sys system with a \emph{multi-stage architecture}.

\bi
\item {\bf The Planning Stage:} Unlike traditional DBMS, where logical operations are already available beforehand via SQL queries, \sys has to determine the appropriate logical data prep operations by analyzing \revise{the semantic alignment between the table and the NL question}.
This process occurs during the planning stage, where the logical plan is formed.
%
%
\item {\bf The Programming Stage:} 
This stage converts the logical plan into a physical plan by generating low-level executable code, selecting the appropriate programming constructs (\eg Python functions or APIs) for each operation, and \revise{customizing the code to align the table’s structure with the NL question's semantics}.
%
%
\item {\bf The Executing Stage:} This stage executes the generated code for each operation and returns any errors encountered to the Programming stage for debugging.
\ei


We implement \sys using the popular LLM-based Multi-Agent framework~\revise{\cite{han2024llm}}, which leverages multiple small, independent agents working collaboratively to solve complex problems.

More specifically, we design a \textsc{Planner} agent, which corresponds to the Planning Stage and suggests a tailored sequence of high-level operations to meet the specific needs of the question, leveraging the semantics understanding and reasoning capabilities of LLMs. The core technical idea behind this \textsc{Planner} agent is a novel {\em Chain-of-Clauses (CoC) reasoning} method. This method translates the NL question into an {\em Analysis Sketch}, which outlines how the table should be prepared to produce the answer, guiding the agent's reasoning based on this sketch. Compared to the popular Chain of Thoughts (CoT) methods~\cite{wei2022chain}, which decompose questions into sub-questions, our approach more effectively captures the semantic relationships between questions and tables.


\sys also includes a set of \textsc{Programmer} agents, each of which synthesizes a question-specific implementation for a given logical data prep operation. However, existing LLM-based code synthesis often generates overly generic code that struggles to effectively address the \emph{heterogeneity} challenges of tables. For instance, values may have diverse syntactic formats (\eg ``\emph{September 1}'' and ``\emph{11-24}'' in $T_6$) or semantic representations (e.g., ``\emph{ITA}'' and ``\emph{Italia}'' in $T_1$), making it difficult to generate code tailored to these variations. To address this, we propose a {\em tool-augmented} approach that enhances the LLM's code generation capabilities by incorporating predefined API functions, which allows the LLM to generate more specialized code that accounts for variations in table values. Furthermore, corresponding to the Executing stage, we design an \textsc{Executor} agent that executes the code to process the table.



\stitle{Contributions.} Our contributions are summarized as follows.

\vspace{1mm} \noindent
(1) We introduce a novel problem of question-aware data preparation for TQA, which is formally defined in Section~\ref{sec:problem}.

\vspace{0.5mm} \noindent
(2) We propose \sys, an LLM-based multi-agent framework for question-aware data prep (Section~\ref{sec:overview}). We develop effective techniques in \sys for the \textsc{Planner} agent (Section~\ref{sec:the_planner_agent}) and the \textsc{Programmer} agents (Section~\ref{sec:the_programmer_and_executor}). 

\vspace{0.5mm} \noindent
(3) We conduct a thorough evaluation on data prep in TQA (Section~\ref{sec:exp}). Extensive experiments show that \sys achieves new SOTA accuracy, outperforming existing TQA methods without data prep by 12.22 points on WikiTQ and 13.23 points on TabFact, and surpassing TQA methods with data prep by 3.05 points on WikiTQ and 1.96 points on TabFact.

\section{Question-Aware Data Prep for TQA}
\label{sec:problem}


\subsection{Tabular Question Answering}
\label{subsec:tqa}

Let $Q$ be a natural language (NL) question, and $T$ a table consisting of $m$ columns (\ie attributes) $\{A_1, A_2, \ldots, A_m\}$ and $n$ rows $\{r_1, r_2, \ldots, r_n\}$, where $v_{ij}$ denotes the value in the $i$-th row and $j$-th column of the table. 
The problem of \textbf{tabular question answering (TQA)} is to generate an answer $Ans$, in response to question $Q$ based on the information in table $T$. By the purposes of the questions, there are two main types of TQA problems: (1) table-based fact verification~\cite{chen2019tabfact, aly2021feverous}, which determines whether $Q$ can be \emph{entailed} or \emph{refuted} by $T$, and (2) table-based question answering~\cite{pasupat2015compositional, nan2022fetaqa}, which extracts or reasons the answer to $Q$ from $T$.
\begin{example}
\label{exam:cases_pre}
	Figure~\ref{fig:issue_instances} provides several examples of TQA. Consider table $T_1$, which contains medal information for cyclists from different countries, with two columns: \term{Cyclist} and \term{Medal}. Given the question $Q_1$, ``Which \textbf{country} has the most medals?'', the answer should be ITA, as two Italian cyclists, ``Dav'' and ``Alex'', have won a total of 4 medals, more than the ESP cyclist ``Alej''. 
    TQA often requires reasoning over tables. For instance, to answer question $Q_2$, we first need to calculate the ``GDP growth rate'' for all countries, then sort the countries by growth rate, and finally identify the country with the highest GDP growth rate, \ie CHN.
\end{example}

\subsection{Data Prep for TQA}
\label{subsec:dataprep-tqa}
%
%
\revise{
In contrast to traditional data prep, \emph{question-aware data prep for TQA} focuses on \emph{adapting the table $T$ to the specific informational needs of a given question $Q$}, thereby enhancing the semantic alignment between the structured table and the NL question.
}


\stitle{Data Prep Operations.}
To meet the new requirements of data prep for TQA, this paper defines high-level, logical \textbf{data prep operations} (or \emph{operations} for short) to formalize the \emph{question-aware} data prep tasks.
Formally, an operation, denoted as $o$, encapsulates a specific question-aware data prep task that transforms table $T$ into another table $T'$, \ie $T' = o(T)$. 

\revise{
As shown in Figure~\ref{fig:error_instance_record}, the majority of TQA errors arise from inadequately addressing three key data prep issues: \emph{missing semantics}, \emph{inconsistent values}, and \emph{irrelevant columns}. To address the challenges, we introduce three types of data prep operations.
}


\bi
	\item \revise{\term{Derive}: 
    A data prep task that derives a new column for table $T$ from existing columns, aimed at addressing the challenge of \emph{missing semantics}. This task typically involves operations such as combining columns through arithmetic computations, extracting relevant values, etc.}
	\item \term{Normalize}: a data prep task that normalizes types or formats of the values in a column of $T$ based on the needs of $Q$, \revise{aimed at addressing the challenge of \emph{inconsistent values}}. This task typically involves value representation or format normalization, type conversion, etc.
	\item \term{Filter}: A data prep task that filters out columns in $T$ that are not relevant to answer question $Q$, \revise{aimed at addressing the challenge of \emph{irrelevant columns}}.   
    This is crucial for handling large tables to address the input token limitations and challenges in long-context understanding of LLMs~\cite{DBLP:journals/corr/abs-2404-02060}.
\ei

Given a high-level operation $o_i$, we define $f_i$ as its low-level \emph{implementation}, either by calling a well-established algorithm from a known Python library or using a customized Python program to meet the requirements of $o_i$. 
%

\begin{example}
	\label{exam:operators}
	Figure~\ref{fig:issue_instances} shows examples of question-aware data prep operations for TQA, along with their implementations in Python.
	
	\vspace{1mm}
    \revise{
	{\em (a) \underline{The \term{Derive} operation:}} Figure~\ref{fig:issue_instances_incomplete} shows three examples of the \term{Derive} operations, \ie extracting \term{Country} information from column \term{Cyclist} in $T_1$, computing new column \term{GrowthRate} using two columns in $T_2$, and inferring a status of \term{IfLost} by analyzing the scores in column \term{Result}. 
	}
    
	\vspace{1mm}
	{\em (b) \underline{The \term{Normalize} operation:}}
	\revise{
    Figure~\ref{fig:issue_instances_inconsistency} shows three examples of \term{Normalize} operations, \ie normalizing the formats of \term{Surface}, \term{Time} and \term{Date} for tables $T_4$, $T_5$ and $T_6$ respectively. 
    }
	
	\vspace{1mm}
	{\em (c) \underline{The \term{Filter} operation:}}
    \revise{
    Figure~\ref{fig:issue_instances_redundancy} illustrates two examples of \term{Filter} operations in tables $T_7$ and $T_8$. 
    The \term{Summary} column in $T_7$ contains an average of 100 words, increasing the difficulty for LLMs to identify the relevant \term{AirDate} column.
    %
    Additionally, although $T_8$ has 12 columns, only two are relevant to $Q_8$, and providing all columns may cause long-context understanding challenges of LLMs~\cite{DBLP:journals/corr/abs-2404-02060}. 
    }
\end{example}
\revise{
\stitle{Remarks.}  
While issues such as missing values and duplicates are common in real-world datasets, they are significantly less prevalent than the three highlighted challenges in existing TQA benchmarks. %
Thanks to \sys's separation of logical operations and physical implementations, the set of data prep operations can be easily extended to accommodate new types of tasks.  
We leave the exploration of broader data prep challenges to future work.
}

\stitle{Question-Aware Data Prep for TQA.}
Given an NL question $Q$ posed over table $T$, \emph{question-aware data prep for TQA} is to generate a sequence of high-level operations $O=\{o_1, o_2, \ldots, o_{|O|}\}$ as a \emph{logical plan}. Then, it generates a \emph{physical plan}, where each operation $o_i$ is implemented by low-level code $f_i$, such that these operations transform $T$ into a new table $T'$ that meets the needs of $Q$.

%

\subsection{\revise{A Straightforward LLM-based Solution}
}
\label{subsec:naive_method_for_question_aware_data_prep}

\begin{figure}[t]
    \centering 
    \includegraphics[width=0.99\columnwidth]{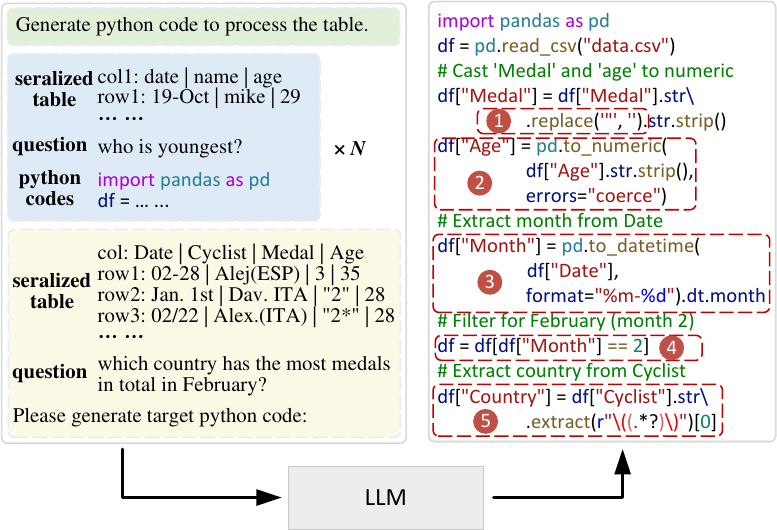}
    \vspace{-1em}
    \caption{An LLM-based method with few-shot prompting for question-aware data prep.
    }
    \label{fig:fewshot_code_based_data_prep}
    \vspace{-1em}
\end{figure}

A straightforward solution to question-aware data prep is to prompt an LLM to prepare tables, leveraging its ability to interpret the specific requirements of NL questions. 
For instance, consider the table in Figure~\ref{fig:fewshot_code_based_data_prep} with columns \term{Date}, \term{Cyclist}, \term{Medal} and \term{Age}, and question: ``\emph{Which country has the most medals in total in February}''. 
A few-shot prompting strategy prompts an LLM with a task description and a few demonstrations, and requests the LLM to generate Python programs as shown in Figure~\ref{fig:fewshot_code_based_data_prep}.

However, this LLM-based solution may encounter the following limitations when performing question-aware data prep for TQA.

\revise{First, at the \emph{logical-operation level}, given the inherent difficulties in understanding both NL questions and tables, it is challenging to accurately identify which data prep operations are specifically required to satisfy the needs of the NL question.  
}
This often leads to false negatives and false positives.  
For example, as shown in Figure~\ref{fig:fewshot_code_based_data_prep}, converting the \term{Age} column to a numerical format in code block \textcircled{2} is a false positive, as it is irrelevant to the question. 
In contrast, failing to normalize the \term{Date} column before extracting the month in code block \textcircled{3} constitutes a false negative, as the method ignores the inconsistency in \term{Date} formats.


Second, at the \emph{physical-operation level}, due to input token limitations and challenges in long-context understanding~\cite{DBLP:journals/corr/abs-2404-02060}, it is not easy to fully understand all possible issues in a table, and thus may struggle to generate customized programs to correct issues.
For example, in Figure~\ref{fig:fewshot_code_based_data_prep}, the normalization of \term{Medal} in code block \textcircled{1} overlooks certain corner cases (\eg ``2*''), and the country extraction in code block \textcircled{5} fails to handle ``Dav.ITA'', which is formatted differently from other values.

Recent methods, such as CoTable~\cite{wang2024chain} and ReAcTable~\cite{zhang2024reactable}, can improve few-shot prompting by employing techniques like Chain-of-Thoughts (CoT) and ReAct. However, these methods remain insufficient to tackle the challenges, as they combine all diverse tasks, such as determining operations and implementing them, within a single LLM agent.
Existing studies~\cite{ishibashi2024self} have shown that a single LLM agent is often ineffective when tasked with handling a diverse range of operations, due to limited context length in LLMs and decreased inference performance with more input tokens.

\begin{figure*}[!t]
    \centering 
    \includegraphics[width=0.95\textwidth]{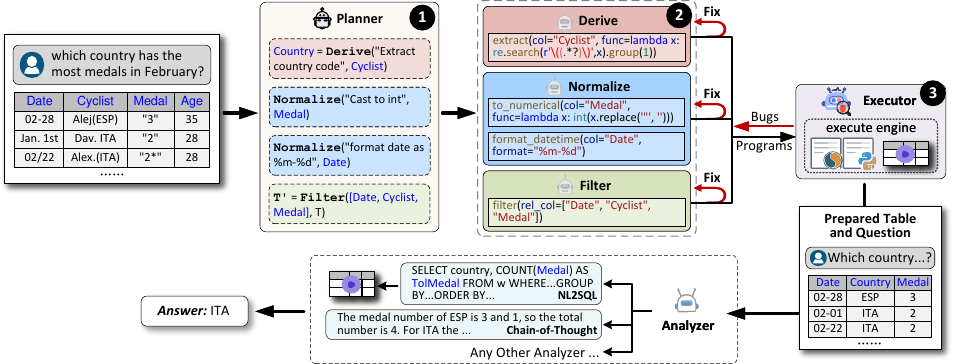}
    \caption{A Multi-agent Data Prep Framework for TQA (Top), which could be plugged to any other TQA methods (Bottom).
    }
    \label{fig:framework_overview}
    \vspace{-1em}
\end{figure*}

\section{An Overview of \sys} 
\label{sec:overview} 

To address the limitations, we propose \sys, a \emph{multi-agent} LLM framework that automatically prepares tables for given NL questions.
Figure~\ref{fig:framework_overview} provides an overview of our framework.
Given an NL question $Q$ posed over table $T$, \sys decomposes the data prep process into three stages:

\be
	\item \textsc{Planner} Agent: the \textbf{Planning} stage. 
    It guides the LLM to suggest \emph{logical} data prep operations $O=\{o_1, o_2, \ldots, o_{|O|}\}$, which are tailored to specific question $Q$, 

	\item Multiple \textsc{Programmer} Agents (\eg \textsc{Normalize}):
    the \textbf{Programming} stage. 
    It directs the LLM to generate \emph{physical} implementation $f_i$ (\eg Python code) for each operation $o_i$ customized for the table $T$. {Besides, it is also tasked for code debugging if any execution errors occur.}

    \item An \textsc{Executor} Agent: the \textbf{Executing} stage. 
    It executes the generated code and {reports errors if any bugs occur.}
\ee


%
%


After that, an \textsc{Analyzer} agent extracts the answer from the prepared table. This agent can either use LLMs as black-boxes or leverage them for code generation, which is \emph{orthogonal} to the question-aware data prep problem studied in this paper. For simplicity, we use a Text-to-SQL strategy that translates the question into an SQL query over the prepared table to obtain the final answer, as shown in Figure~\ref{fig:framework_overview}. Note that other strategies could also be used by the agent in a ``plug-and-play'' manner.

\begin{example}
	Figure~\ref{fig:framework_overview} illustrates how \sys supports data prep for an NL question
    posed over a table with $4$ columns.
	
	\vspace{1mm}
	{\em (a) \underline{The \term{Planning} stage:}} The \textsc{Planner} suggests the following high-level operations to address the specific NL question: 
	\bi
		\item \term{$T'$[Country]=Derive(}``Extract country code'', \term{Cyclist)} that extracts the country information from column \term{Cyclist}, producing a new \term{Country} column, in response to the ``which country'' part of the question. 
		\item \term{Normalize(}``Case to INT'', \term{Medal)} that standardizes the value formats in the \term{Medal} column (\eg removing quotation marks and asterisks) and then converts the strings to integers, as the question requires ``the most medals'';
		\item \term{Normalize(}``Format date as \%m-\%d'', \term{Date)} standardizes the values in the \term{Date} column into a unified format to support the ``in February'' condition in the question.
		\item $T'$ $=$ \term{Filter(}\term{[}\term{Date}, \term{Country}, \term{Medal]}, $T$\term{)} that filters out column \term{Age}, which is irrelevant to the question; 
	\ei
	
	\vspace{1mm}
	{\em (b) \underline{The \term{Programming} stage:}} \sys designs specialized \textsc{Programmer} agents for each operation type, \ie \textsc{Derive}, \textsc{Normalize} and \textsc{Filter}.
	Each specialized \textsc{Programmer} focuses on generating executable code for its assigned operations. 

 	\vspace{1mm}
	{\em (c) \underline{The \term{Executing} stage:}} an Executor agent iteratively refines the generated code if any error occurs.
	
	After these stages, \sys generates a prepared table $T^*$, which is then fed into an \textsc{Analyzer} agent to produce the answer \term{ITA}. 
\end{example}



\stitle{The \textsc{Planner} Agent.}
The key challenge is how to suggest a \emph{logical plan} that address specific NL questions. 
Even for the same table, different NL questions may require not only different logical operations but also varying sequences of those operations. 
To address this challenge, we propose a novel {\em Chain-of-Clauses (CoC) reasoning} method for the \textsc{Planner} agent. This method translates the NL question into an {\em Analysis Sketch}, outlining how the table should be transformed to produce the answer, thereby guiding the agent's reasoning based on this sketch. 
More details of the method are given in Section~\ref{sec:the_planner_agent}. 

\stitle{The \textsc{Programmer} Agents.}
The key challenge is that a given logical operation can have multiple executable code alternatives (\eg Python functions), and the difference in outcomes between the best and worst options can be substantial. For example, the \textsc{Derive} agent may generate an overly generic regular expression that extracts countries based on parentheses. Unfortunately, this code fails to correctly process ``Dav. ITA'', which is formatted differently from other values.
To tackle this challenge, we develop a {\em tool-augmented} approach that enhances the LLM's code generation capabilities by utilizing predefined API functions. 
More details of our tool-augmented approach are discussed in Section~\ref{sec:the_programmer_and_executor}.


\revise{
\stitle{Remarks.}
Our proposed \sys framework is extensible. When additional question-aware data prep operations are required, more specialized \textsc{Programmer} agents can be designed to handle them. The central \textsc{Planner} agent can then determine which operations should be performed and assign them accordingly.
}

\section{The Planner} 
\label{sec:the_planner_agent} 

%


\subsection{A Direct Prompting Method} \label{subsec:planner-prompting}
The most common way to generate a logical plan is to directly prompt an LLM using a typical in-context learning approach.
The inputs are a question $Q$, a table $T$, a set $\Sigma$ of specifications for each operation type, and an LLM $\theta$. Here, each specification $\sigma \in \Sigma$ describes the purpose of an operation type, e.g., ``\emph{an \term{Derive} operation creates a new column for a table based on existing columns, in response to the specific needs of a question}''.
The output of the algorithm is a set $O$ of high-level operations

%

\begin{figure*}[!t]
    \centering 
    \includegraphics[width=0.95\textwidth]{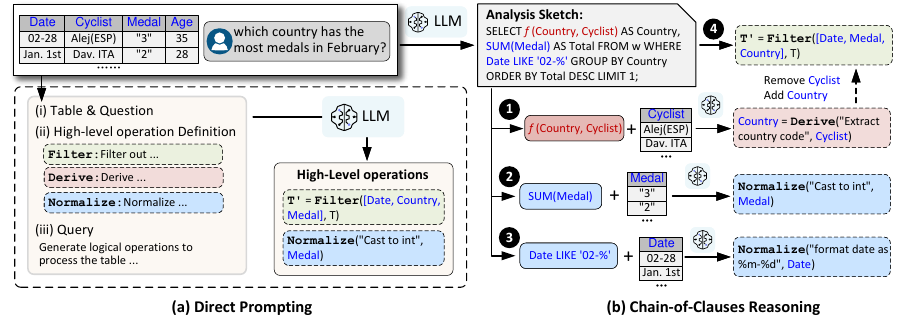}
    \vspace{-1em}
    \caption{Our proposed approaches to logical plan generation in the \textsc{Planner} agent: (a) A straightforward direct prompting method, and (b) a more effective Chain-of-Clauses (CoC) reasoning method.
    }
    \label{fig:planner-cot}
    \vspace{-1em}
\end{figure*}

\begin{example}
	Figure~\ref{fig:planner-cot}(a) illustrates the direct prompting method, which produces two logical operations, \term{Filter} and \term{Normalize}. However, this logical plan might not be accurate, as discussed below. 
	
	\vspace{1mm}
	{\em (a) \underline{Incorrect operations:}} 
    \revise{
    The \term{Filter} operation retains the \term{Country} column simply because the question mentions ``which country.'' However, it fails to recognize that the original table does not actually contain a \term{Country} column. Worse yet, it incorrectly filters out the \term{Cyclist} column, merely because \term{Cyclist} is not explicitly mentioned in the question. This mistake is critical, as the country information is implicitly embedded within the \term{Cyclist} values.
    }
	
	\vspace{1mm}
	{\em (b) \underline{Missing operations:}} Observing the ground-truth in Figure~\ref{fig:framework_overview}, we can see that the \term{Derive} operation on \term{Cyclist} is not generated, as the column has already been filtered out. Moreover, although the \term{Normalize} operation on \term{Medal} is generated, the \term{Normalize} operation on \term{Date} is missing. This is because the phrase ``the most'' in the question suggests the need for type conversion of \term{Medal}, but there are no clues from the question to normalize \term{Date}, unless we observe from the table that the values are inconsistent.
    
\end{example}

%

The above example clearly demonstrates that the key challenge in designing the \textsc{Planner} agent, as discussed earlier, is to capture \emph{the relationships between different parts of question $Q$ and the columns in table $T$}. For instance, as shown in Figure~\ref{fig:framework_overview}, the \term{Normalize} operation on the \term{Date} column is generated by considering two key factors: (1) the question contains ``in February'', and (2) the values in the \term{Date} column are inconsistent.

\subsection{A Chain-of-Clauses Method } \label{subsec:planner-cot}

To address this challenge, we propose a {\em Chain-of-Clauses (CoC) reasoning} method that decomposes the entire process of logical plan generation into two phases, as shown in Figure~\ref{fig:planner-cot}(b). 
\bi
	\item The first phase leverages an LLM to generate an SQL-like {\em Analysis Sketch}, outlining how the table should be transformed to produce the answer. 
	\item The second phase iteratively examines different clauses in the Analysis Sketch, \eg~\term{ORDER} \term{BY} \term{Date}.
    \revise{At first, we use the syntax parser from Binder~\cite{cheng2023binding} to extract all UDFs and operation clauses in the sketch and sort them by the execution order.
    Then, }for each clause, we associate the corresponding data in $T$ (\eg values in column \term{Date}) with it {and provide several in-context learning demos} to prompt the LLM for generating possible logical operations (\eg~\term{Normalize}). 
\ei

Compared with existing CoT methods~\cite{wei2022chain}, which simply break down questions into sub-questions, our approach is more effective for logical plan generation. First, our method decomposes the question into a set of analysis steps over table, formalized as an SQL-like Analysis Sketch, which simplifies the task of logical plan generation. More importantly, our method \emph{jointly} considers each analysis step along with the corresponding relevant columns (instead of the whole table) to prompt the LLM, effectively capturing the relationships between the question and the data.


\stitle{SQL-like Analysis Sketch.}
An Analysis Sketch $S$ is an SQL-like statement, which can be represented as follows. 
\begin{lstlisting}[
	mathescape=true,
	language=SQL,
	showspaces=false,
	basicstyle=\ttfamily,
	numbers=none,
	xleftmargin=0cm,
	xrightmargin=0cm,
	numberstyle=\tiny,
	commentstyle=\color{gray}
	]
	SELECT $A~|~{\tt agg}(A)~|~f(A^*_i, \{A_i\})$ FROM $T$
	WHERE ${\tt Pred}(A_i)$ AND $\ldots$ AND ${\tt Pred}(A_j)$
	GROUP BY $A$ ORDER BY $A$ LIMIT $n$
\end{lstlisting}
where ${\tt agg}(A)$ is an aggregation function (\eg \term{SUM} and \term{AVG}) over column $A$, ${\tt Pred}(A_i)$ denotes a predicate (\ie filtering condition) over column $A_i$, such as \term{Date} \term{LIKE} '\term{02-\%}. 
Note that $f(A^*_i, \{A_i\})$ is a \emph{user-defined function} (UDF) that maps existing columns $\{A_i\}$ in the table into a new column $A^*_i$. 
Figure~\ref{fig:planner-cot}(b) shows an example Analysis Sketch with a UDF that specifies a new column \term{Country}, \ie $f(\term{Country}, \term{Cyclist})$. Obviously, this UDF is introduced because the Analysis Sketch contains a \term{GROUP} \term{BY} \term{Country} clause.

\revise{
\stitle{Logical Operation Generation.}
Given the constructed Analysis Sketch, we orchestrate the logical operations and determine their execution order by translating the sketch into actionable steps.
Specifically, we leverage the syntax parser from Binder~\cite{cheng2023binding} to extract all UDFs and operation clauses from the sketch and sort them according to their intended execution sequence from the parser.
Based on this parsed order, we then generate the corresponding logical operations, as illustrated below.
}

%
\begin{example}\label{example:cot}
	Figure~\ref{fig:planner-cot}(b) shows our proposed CoC reasoning method for the example question and table in Figure~\ref{fig:framework_overview}. Specifically, the method generate a set $O$ of operations via the following two phases. 
	
	\vspace{1mm}
	{\em (a) \underline{Phase I - Analysis Sketch Generation:}} 
	In this phase, the algorithm prompts the LLM $\theta$ with several exemplars $\{(Q_i, T_i, s_i)\}$ to generate an Analysis Sketch $S$, as shown in Figure~\ref{fig:planner-cot}(b). 
	
	\vspace{1mm}
	{\em (b) \underline{Phase II - Operation Generation:}} 
	In this phase, the algorithm iteratively examines the clauses in Analysis Sketch $S$ as follows. 
	(1) $f(\term{Country, Cyclist})$: this clause and relevant columns are used to prompt the LLM to generate an \term{Derive} operation.
    (2) ${\tt SUM}(\term{Medal})$: this clause and the values in column \term{Medal} are used to prompt the LLM to generate a \term{Normalize} operation.
    (3) \term{Date} \term{LIKE} `\term{02-\%}': this clause and the values in column \term{Date} are used to prompt the LLM to generate a \term{Normalize}.
	Finally, the algorithm generates a \term{Filter} operation that only selects columns relevant to the Analysis Sketch. 
    \revise{Note that the \term{Filter} operation removes irrelevant columns, whereas the \term{Derive} operation creates new columns from existing ones.}
	 
\end{example}

\section{The Programmer \& Executor} 
\label{sec:the_programmer_and_executor}

\textsc{Programmer} agents translate a high-level logical plan into a \emph{physical plan} by generating low-level code, which is then passed to an \textsc{Executor} agent for code execution and interactive debugging. A straightforward approach is to prompt an LLM with the logical plan using in-context learning with several exemplars, asking the LLM to generate code for each high-level operation in the plan. The code is then executed iteratively, and if any runtime errors occur, the \textsc{Programmer} is prompted with the error messages for debugging.

However, the above method may have limitations, as the generated code may be overly generic and unable to effectively address the \emph{heterogeneity} challenge of the tables. Specifically, many tables originate from web sources or real-world scenarios, where values have diverse syntactic (\eg ``19-Oct'' and ``9/14'') or semantic formats (\eg ``ITA'' and ``Italia''), making it difficult to generate code customized to these tables.
For example, 
given a \term{Normalize} operation to standardize country formats, a generic function from an existing Python library may not be sufficient to transform country names (\eg ``Italia'') to their ISO codes (\eg ``ITA''). In such cases, customized functions, such as the \term{clean\_country} function from an external library \term{dataprep}~\cite{DBLP:conf/sigmod/PengWLBYXCRW21}, are needed to address specific requirements of code generation.

To address these limitations, we introduce a \emph{tool-augmented} method for the \textsc{Programmer} agents, enhancing the LLM's code generation capabilities by utilizing pre-defined API functions, referred to as the \emph{function pool} in this paper. 


\subsection{Tool-Augmented Method: Key Idea}
\label{subsec:prog-idea}

\begin{figure*}[t]
    \centering 
    \includegraphics[width=0.95\textwidth]{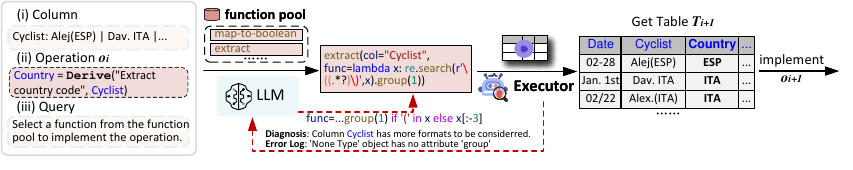}
    \vspace{-1em}
    \caption{{Our proposed tool-augmented method for physical plan generation and execution.}}
    \label{fig:tool_augmented_method}
    \vspace{-1em}
\end{figure*}

Figure~\ref{fig:tool_augmented_method} provides an overview of our \emph{tool-augmented} method for physical plan generation and execution. Given a table $T$ and a set $O=\{o_1, \ldots, o_{|O|}\}$ of logical operations, the method \emph{iteratively} generates and executes physical executable code for each individual operation $o_i \in O$, thus generating a sequence of intermediate tables $T_1$ (\ie the input $T$), $T_2$, \ldots $T_{|O|+1}$.  
Specifically, in the $i$-th iteration, given a high-level operation $o_i$ (\eg~\term{Derive}) and an intermediate table $T_i$, the method generates physical code and executes it to produce an updated table $T_{i+1}$ through the following two steps. 

\stitle{Function Selection and Argument Inference.}
The method first prompts the LLM to select a specific function from a function pool $\mathcal{F}$ corresponding to the operation type (e.g., $F_{\tt Der}$) and infer the arguments (\eg regular expression) for the selected function.

For instance, given the \term{Derive} operation over table $T_i$ shown in Figure~\ref{fig:tool_augmented_method}, the LLM selects a function from the pool $\mathcal{F}_{\tt Der}$ designed specifically for \term{Derive}, obtaining an \term{extract} function with two arguments: \term{column} and \term{func}. The LLM then generates preliminary code for these arguments, assigning \term{Cyclist} to the \term{column} argument and generating a lambda function with a specific regular expression for argument \term{func}.

\revise{
Note that, in addition to selecting functions from the corresponding pool, our method may also prompt the LLM to write specific code for an operation (denoted as $\mathtt{CustomCode}$) if no existing function is suitable to meet the operation's requirements.
}

\stitle{Code Execution and Debugging.}
Given the function generated in the previous step, the \textsc{Executor} agent then applies the function over table $T_i$. {If any bugs occur during execution, it captures and summarizes error messages and returns to corresponding \textsc{Programmer} agent.}
%
For instance, as shown in Figure~\ref{fig:tool_augmented_method}, the first lambda function for the argument \term{func} only extracts countries based on parentheses, which may produce incorrect results for the value ``Dav.ITA'', as it is formatted differently from other values. In this case, based on the execution results, the \textsc{Executor} agent {records the error log and examines relevant data to summarize reasons, which will be passed to the corresponding \textsc{Programmer} to modify the code.}
%
%
After {the execution and debugging process,}
the method produces a new intermediate table $T_{i+1}$ and proceeds to the next operation, $o_{i+1}$, \ie a \term{Normalize} operation in Figure~\ref{fig:tool_augmented_method}.

\subsection{Search Space of Function Pools}
\label{subsec:prog-pool}

This section presents a search space of function pools for different types of high-level operations formalized in the current \sys. 

\stitle{Function Pool for $\mathtt{Derive}$.} We define a specific \textsc{Programmer} agent, \ie~\textsc{Derive}, to generate low-level code for the operation type $\mathtt{Derive}$ from the following function pool $\mathcal{F}_{\mathtt{Der}}$.

(1) $\mathtt{extract}$.
	This function extracts a substring from a column to generate a new column. It has two arguments: ${\tt column}$, representing the name of the source column, and ${\tt func}$, representing the substring extraction function. Using table $T_1$ from Figure~\ref{fig:issue_instances_incomplete} as an example, to extract the country code, we would set ${\tt column}$ to \term{Cyclist} and ${\tt func}$ to a lambda function {\small\texttt{lambda x: re.search(r'\((.*?)\)',x).group(1)}}.
    
(2) $\mathtt{calculate}$.
	The function generates a new column through arithmetic operations of existing columns.
    It has two arguments: ${\tt columns}$, representing a list of source columns and ${\tt func}$, representing a lambda function with input to be a dictionary of values, since we may perform calculations on multiple columns.
    For example, to generate a new column \term{GrowthRate} for $T_2$ in Figure~\ref{fig:issue_instances_incomplete}, we set ${\tt columns}$ to {\small \texttt{[`2012', `2013']}} and ${\tt func}$ to a lambda function {\small \texttt{lambda x: (x[`2013']-x[`2012'])/x[`2012']}}.
	
(3) $\mathtt{map\mbox{-}to\mbox{-}boolean}$ and $\mathtt{concatenate}$. Both functions generate a new column from multiple columns. The difference is that one generates boolean values, while the other concatenates strings. Both operators take two arguments, ${\tt columns}$ and ${\tt func}$, similar to those in the $\mathtt{calculate}$ operator.
	
(4) $\mathtt{infer}$.
	This function uses LLMs to deduce values that could not be processed by the above functions. It takes a list of columns and the failed target values as input, learning from demonstrations of successfully processed values to output the target values.

\stitle{Function Pool for $\mathtt{Normalize}$.} 
We define a specific \textsc{Programmer} agent, \ie \textsc{Normalize}, to generate low-level code for the operation type $\mathtt{Normalize}$ from the following function pool $\mathcal{F}_{\mathtt{Norm}}$.
%

(1) $\mathtt{to\mbox{-}numerical}$. 
	This function standardizes a column to a numerical type, such as int or float. To use it, two arguments need to be specified: ${\tt column}$, representing the source column, and ${\tt func}$, representing a lambda function. For example, 
    we can use this function to standardize the \term{Time} column to integers, with ${\tt column}$ as \term{Time} and ${\tt func}$ as
    the lambda function.

    (2) $\mathtt{format\mbox{-}datetime}$. 
	The function standardizes the format of columns with \term{DateTime} values. It has two arguments: ${\tt column}$ and ${\tt format}$. The ${\tt format}$ argument specifies the desired format for standardization. 
    For example, 
    to make the values in the \term{Date} column comparable, we use the $\mathtt{format\mbox{-}datetime}$ function with ${\tt column}$ set to \term{Date} and ${\tt format}$ set to ``\%Y-\%m-\%d''.
    
(3) $\mathtt{clean\mbox{-}string}$. To clean string values in tables, we define function $\mathtt{clean\mbox{-}string}$ with two arguments: ${\tt column}$ and ${\tt trans\_dict}$. The $\tt trans\_dict$ is a dictionary where each key $k$ and its corresponding value $v$ represent that $k$ in $\tt column$ is replaced by $v$. 

(4) $\mathtt{infer}$. This function uses LLMs to deduce values that could not be processed by the functions above.

\stitle{Function Pool for \term{Filter}.} 
We design the \textsc{Filter} agent to utilize a function pool $\mathcal{F}_{\mathtt{Filter}}$ with a single function, $\mathtt{filter\mbox{-}columns}$. This function has one argument, ${\tt rel\_columns}$, which represents a list of question-related column names.

\revise{
\stitle{Remarks.}  
The design of $\mathtt{CustomCode}$ and $\mathtt{infer}$ operations enables \sys to generalize to unseen tasks.  
For example, when encountering missing or clearly erroneous values, \sys can directly prompt an LLM to infer plausible replacements.
Moreover, our search space is extensible: additional functions or external APIs can be easily integrated into the function pool, including specialized data preparation functions for various semantic types, such as those provided by the \term{dataprep} library~\cite{DBLP:conf/sigmod/PengWLBYXCRW21}.
%
%
}

\section{Experiments}
\label{sec:exp}






\subsection{Experimental Setup} 
\label{subsec:exp-setup}

\stitle{Datasets.} 
We evaluate \sys using the following well-adopted datasets.
The statistics of the datasets are shown in Table~\ref{tbl:datasets}.


\vspace{1mm}
\noindent
(1) \textbf{WikiTQ}~\cite{pasupat2015compositional} contains complex questions annotated by crowd workers based on diverse Wikipedia tables. WikiTQ comprises 17,689 question-answer pairs in the training set and 4,344 in the test set with significant variation in table size. 

\vspace{1mm}
\noindent 
(2) \textbf{TabFact}~\cite{chen2019tabfact} provides a collection of Wikipedia tables, along with manually annotated NL statements. One requires to deduce the relations between statements and tables as ``true'' (a statement is entailed by a table) or ``false'' (a statement if refuted by a table). 
    To reduce the experimental cost without losing generality, we choose the small test set provided by~\cite{ye2023large} which contains 2024 table-statement pairs.
    The tables in TabFact are much smaller than those in WikiTQ. The answers are binary with nearly equal proportions.

\vspace{1mm}
\noindent 
(3) \textbf{TabBench}~\cite{wu2024tablebench} provides an evaluation of TQA capabilities within four major categories, including multi-hop fact checking (FC), multi-hop Numerical Reasoning (NR), Trend Forecasting and Chart Generation. We evaluate the generalization of \sys on FC and NR, where one requires to conduct more complex and multi-hop reasoning over tables for answering a question. \revise{Specifically, we directly use the prompts in WikiTQ to evaluate the generalization capabilities of \sys on TabBench.}

{\stitle{Baselines.}} {
We consider the following three categories of baselines:

\vspace{1mm} \noindent
{(1) {\bf TQA Methods w/o Data Prep}} directly generate the answer or utilize programs to extract the answer from the table without considering data prep. We implement four primary TQA baselines.

\etitle{\textbf{End2EndQA (End2End)}~\cite{chen2022large}} utilizes the in-context learning abilities of LLMs to generate the answer for TQA task based on the supervision of human-designed demonstrations. We implement End2End method with prompt and demonstrations provided by~\cite{cheng2023binding}.

\etitle{\textbf{Chain-of-Thought (CoT)}~\cite{wei2022chain}} prompts LLMs to generate the reasoning process step-by-step before generating the final answer. We implement CoT with the prompt provided by~\cite{chen2022large}.

\etitle{\textbf{NL2SQL}~\cite{rajkumar2022evaluating}} first translates the question into an SQL program and then executes it to get the final answer from the table for the question. We use the prompt from~\cite{cheng2023binding} to implement NL2SQL.

\etitle{\textbf{NL2Py}} uses Python code to process and reason over the tables. To construct the prompt for NL2Py, we use TQA instances in NL2SQL prompt and manually write the Python code to process the table and generate the final answer.

\begin{table}[t]
\centering
\caption{\bf{Statistics of Datasets.}} 
\renewcommand{\arraystretch}{1.14}
  \vspace{-0.5em}
   \resizebox{0.95\linewidth}{!}{
\begin{tabular}{|c|c|c|c|c|}
\hline
\textbf{Dataset}  & \textbf{\# Rec.}     & \textbf{\# Row.} & \textbf{\# Col.} & \textbf{Ans. Types} \\ 
\hline \hline
WikiTQ    & $22,033$       & 4$\sim$753   & 3$\sim$25  &  string / list (3.05\%)    \\ \hline
TabFact  & $2,024$       & 5$\sim$47   & 5$\sim$14  &  true / false (49.60\%)      \\ \hline
{TableBench}  & {$886$}       & {2$\sim$212}   & {2$\sim$20}  &  {string / list (31.49\%)}      \\ \hline
\end{tabular}
}
\label{tbl:datasets}
  \vspace{-1em}
\end{table}

\revise{
\vspace{1mm} \noindent
{(2) {\bf Data Prep Baselines}}. 
We consider the following data prep baselines. For each baseline, we first perform data prep and then use the above NL2SQL to extract answers from the prepared tables.

\etitle{\textbf{Offline DataPrep (Off-Prep)}}
performs offline data prep operations for all tables in TQA.
To this end, we have surveyed and consolidated the offline data prep operations, such as data cleaning, value normalization and column renaming, adopted by current SOTA TQA methods~\cite{yao2022react, cheng2023binding, khot2022decomposed, wang2024chain, DBLP:journals/pvldb/ZhuCXLSZSTL24} to construct a comprehensive offline data prep pipeline, by utilizing the Pandas library~\cite{mckinney-proc-scipy-2010} and the popular DataPrep toolkit~\cite{dataprepeda2021}.


\etitle{\textbf{ICL-Prep}} uses few-shot ICL demonstrations to guide LLMs in generating Python programs for data prep, as shown in Figure~\ref{fig:fewshot_code_based_data_prep}.
}

\vspace{1mm} \noindent
{(3) {\bf SOTA TQA Methods with Data Prep}}. We investigate four SOTA TQA methods considering data prep tasks for TQA task.

\etitle{\textbf{Dater~\cite{ye2023large}}} addresses the TQA task by decomposing the table and question. It first selects relevant columns and rows to obtain a sub-table and then decomposes the origin question into sub-questions. Dater answers these sub-questions based on the sub-tables to generate the final answer. We use code in~\cite{codeofdater} for implementation.

\etitle{\textbf{Binder~\cite{cheng2023binding}}} enhances the NL2SQL method by integrating LLMs into SQL programs. It uses LLMs to incorporate external knowledge bases and directly answer questions that are difficult to resolve using SQL alone. We utilize the original code provided by~\cite{codeofbinder}.

\revise{
\etitle{\textbf{AutoTQA~\cite{DBLP:journals/pvldb/ZhuCXLSZSTL24}}} uses a multi-agent framework for TQA.
Since the official code for AutoTQA is not publicly available, we reimplement it under the guidance of the authors (see our repository ({https://github.com/fmh1art/AutoPrep/src/model/autotqa}).
Given the high time and API costs, we evaluate AutoTQA on a sampled subset of 500 instances from each of WikiTQ and TabFact.
}

\etitle{\textbf{ReAcTable~\cite{zhang2024reactable}}} uses the ReAct paradigm to extract relevant data from the table using Python or SQL code generated by LLMs. Once all relevant data is gathered, it asks the LLMs to predict the answer. We run the original code from~\cite{codeofreactable} and keep all settings as default. Notice that the original code does not include prompts for TabFact, we generate it based the WikiTQ prompt.

\etitle{\textbf{Chain-of-Table (CoTable)~\cite{wang2024chain}}} enhances the table reasoning capabilities of LLMs by predefining several common atomic operations (including data prep operations) that can be dynamically selected by the LLM. These operations form an ``operation chain'' that represents the reasoning process over a table and can be executed either via Python code or by prompting the LLM. We implement CoTable using the original code from~\cite{codeofcotable}. 

\begin{table*}[t]
  \centering
  \caption{\bf{
  Improvement of data prep for TQA (the best results are in bold and the second-best are underlined).
  }
  }
  \vspace{-0.5em}
\resizebox{0.9\linewidth}{!}{
    \begin{tabular}{|c||c|c|c|c|c|c|c|c|}
      \hline
      \multirow{2}{*}{\textbf{Method}} & \multicolumn{2}{c|}{\textbf{DeepSeek}} & \multicolumn{2}{c|}{\textbf{GPT3.5}} & \multicolumn{2}{c|}{\textbf{Llama3}} & \multicolumn{2}{c|}{\textbf{QWen2.5}} \\ \cline{2-9}
                   & \textbf{WikiTQ}                                                       & \textbf{TabFact}                                                       & \textbf{WikiTQ}                                                       & \textbf{TabFact}                                                        & \textbf{WikiTQ}                                                        & \textbf{TabFact}                                             & \textbf{WikiTQ}                                               & \textbf{TabFact}                                             \\
      \hline \hline
      End2End      & ${56.65}$                                                             & ${81.77}$                                                              & $52.56$                                                               & ${71.54}$                                                               & ${58.72}$                                                              & ${81.27}$                                                    & ${60.01}$                                                     & ${81.17}$                                                    \\ \cline{1-9} 
      {~~~~+ \sys}     & $63.14$ {\scriptsize \textcolor{cadmiumgreen}{$\uparrow6.49$}}             & $82.11$ {\scriptsize \textcolor{cadmiumgreen}{$\uparrow0.34$}}              & $61.21$ {\scriptsize \textcolor{cadmiumgreen}{$\uparrow8.65$}}             & $71.79$ {\scriptsize \textcolor{cadmiumgreen}{$\uparrow0.25$}}               & $61.23$ {\scriptsize \textcolor{cadmiumgreen}{$\uparrow2.51$}}              & $84.19$ {\scriptsize \textcolor{cadmiumgreen}{$\uparrow2.92$}}    & $63.42$ {\scriptsize \textcolor{cadmiumgreen}{$\uparrow3.41$}}     & $83.05$ {\scriptsize \textcolor{cadmiumgreen}{$\uparrow1.88$}}    \\ \cline{1-9} 
      \hline \hline
      CoT          & $54.95$                                                               & ${82.02}$                                                              & ${53.48}$                                                             & $65.37$                                                                 & $40.75$                                                                & ${80.93}$                                                    & ${59.67}$                                                     & ${82.31}$                                                    \\ \cline{1-9} 
      {+ \sys    } & $61.12$  {\scriptsize \textcolor{cadmiumgreen}{$\uparrow6.17$}}            & $82.26$ {\scriptsize \textcolor{cadmiumgreen}{$\uparrow0.24$}}              & $60.01$ {\scriptsize \textcolor{cadmiumgreen}{$\uparrow6.53$}}             & $74.36$ {\scriptsize \textcolor{cadmiumgreen}{$\uparrow8.99$}}               & $56.01$ {\scriptsize \textcolor{cadmiumgreen}{$\uparrow15.26$}}             & $83.65$ {\scriptsize \textcolor{cadmiumgreen}{$\uparrow2.72$}}    & $62.02$ {\scriptsize \textcolor{cadmiumgreen}{$\uparrow2.35$}}     & $\underline{85.67}$ {\scriptsize \textcolor{cadmiumgreen}{$\uparrow3.36$}}  \\ \cline{1-9} 
      \hline \hline
      NL2Py      & ${59.35}$                                                             & $68.13$                                                                & ${53.59}$                                                             & ${66.15}$                                                               & $50.12$                                                                & $76.24$                                                      & $53.02$                                                       & $72.63$                                                      \\ \cline{1-9} 
      {+ \sys}     & $\underline{65.86}$ {\scriptsize \textcolor{cadmiumgreen}{$\uparrow6.51$}} & $\underline{87.35}$ {\scriptsize \textcolor{cadmiumgreen}{$\uparrow19.22$}} & $\underline{64.69}$ {\scriptsize \textcolor{cadmiumgreen}{$\uparrow11.1$}} & $\underline{84.83}$  {\scriptsize \textcolor{cadmiumgreen}{$\uparrow18.68$}} & $\underline{62.55}$ {\scriptsize \textcolor{cadmiumgreen}{$\uparrow12.43$}} & $\underline{85.42}$ {\scriptsize \textcolor{cadmiumgreen}{$\uparrow9.18$}}  & $\underline{68.65}$ {\scriptsize \textcolor{cadmiumgreen}{$\uparrow15.63$}}  & $\textbf{85.72}$ {\scriptsize \textcolor{cadmiumgreen}{$\uparrow13.09$}} \\ \cline{1-9} 
      \hline \hline
      NL2SQL       & $52.83$                                                               & $70.21$                                                                & $52.90$                                                               & $64.71$                                                                 & ${51.80}$                                                              & $75.15$                                                      & $56.86$                                                       & $80.09$                                                      \\ \cline{1-9} 
      {+ \sys }    & $\textbf{66.09}$ {\scriptsize \textcolor{cadmiumgreen}{$\uparrow13.26$}}   & $\textbf{87.85}$ {\scriptsize \textcolor{cadmiumgreen}{$\uparrow17.64$}}    & $\textbf{64.75}$ {\scriptsize \textcolor{cadmiumgreen}{$\uparrow11.85$}}   & $\textbf{84.19}$ {\scriptsize \textcolor{cadmiumgreen}{$\uparrow19.48$}}     & $\textbf{63.72}$ {\scriptsize \textcolor{cadmiumgreen}{$\uparrow11.92$}}    & $\textbf{85.72}$ {\scriptsize \textcolor{cadmiumgreen}{$\uparrow10.57$}} & $\textbf{68.72}$  {\scriptsize \textcolor{cadmiumgreen}{$\uparrow11.86$}} & $85.33$ {\scriptsize \textcolor{cadmiumgreen}{$\uparrow5.24$}}    \\
      \hline
    \end{tabular}
  }
  \label{tbl:improve_on_base_TQA_method}
  \vspace{-1em}
\end{table*}

\revise{
\stitle{Evaluation Metrics.} We consider both accuracy and cost.

\etitle{\bf Accuracy.}
We adopt the evaluator from Binder~\cite{cheng2023binding} to address cases where program executions are semantically correct but do not exactly match the golden answers.
%
%

\etitle{\bf Cost.}
We measure both the time and API cost for all methods.
For time cost, we ensure a stable network environment and record the end-to-end processing time for each method on a single TQA instance.
For API cost, we follow the official pricing guidelines ({https://api-docs.deepseek.com/quick\_start/pricing/}) and calculate the cost based on the default LLM backbones used.
%
}

{\stitle{Backbone LLMs}}. We evaluate our methods using representative LLMs as backbones.
For closed-source LLMs, we select DeepSeek~\cite{DBLP:journals/corr/abs-2401-14196} (DeepSeek-V2.5-Chat) and GPT3.5~\cite{DBLP:conf/nips/BrownMRSKDNSSAA20} (GPT3.5-Turbo-0613).
For open-source LLMs, we choose Llama3~\cite{dubey2024llama} (Llama-3.1-70B-Instruct) and QWen2.5~\cite{yang2024qwen2} (QWen2.5-72B-Instruct) for evaluation.

\stitle{Experiment Settings.}
\revise{
We provide {detailed prompts} of each component in \sys in our {technical report}~\cite{technical_report} due to the space limit.
}
Moreover, for fair comparison, we set the maximum token input of all methods as 8192. Moreover, we set the temperature parameter of all methods to 0.01 for reproducibility.
}

\subsection{Improvement of Data Prep for TQA}
\label{subsec:Improvement_of_Data_Prep_for_TQA}

\noindent
\textbf{Exp-\expnum\label{exp:improve_by_autoprep}: Impact of question-aware data prep on TQA performance.}
We integrate \sys into our four TQA baselines w/o data prep, 
and report the results in Table~\ref{tbl:improve_on_base_TQA_method}. 


As demonstrated, integrating \sys significantly improves the performance of all evaluated methods. Notably, NL2SQL shows the most substantial gains, achieving an average accuracy improvement of \textbf{12.22} on WikiTQ and \textbf{13.23} on TabFact across all LLM backbones.
Similarly, NL2Py also shows notable improvements in its performance, after being integrated with \sys.  
This significant improvement is attributed to the sensitivity of NL2SQL and NL2Py to data incompleteness and inconsistency, which can lead to erroneous outcomes when performing operations on improperly formatted data. Thus, data prep operations, such as \term{Derive} and \term{Normalize} can solve these cases and improve the overall results.

Moreover, End2End and CoT methods also show considerable performance gains. 
These improvements are largely due to the filtering mechanism of \sys, which removes irrelevant columns, thereby simplifying the reasoning process for extracting answers from tables.
Since ``NL2SQL + \sys'' achieves the best accuracy in most cases, we take its results as default for further comparison.

%


\subsection{Data Prep Method Comparison}
\label{subsec:Data_Prep_Method_Comparision}

\begin{table*}[t]
  \centering
  \caption{\bf{\revise{Experimental results of \sys and TQA methods with data prep.}}
  }
  \vspace{-0.5em}
\resizebox{0.8\linewidth}{!}{
    \begin{tabular}{|c||c|c|c|c|c|c|c|c|}
      \hline
      \multirow{2}{*}{\textbf{Method}} & \multicolumn{2}{c|}{\textbf{DeepSeek}} & \multicolumn{2}{c|}{\textbf{GPT3.5}} & \multicolumn{2}{c|}{\textbf{Llama3}} & \multicolumn{2}{c|}{\textbf{QWen2.5}} \\ \cline{2-9}
                                                      & \textbf{~WikiTQ~}                                                       & \textbf{~TabFact~}                                                       & \textbf{~WikiTQ~}                                                       & \textbf{~TabFact~}                                                       & \textbf{~WikiTQ~}                                                        & \textbf{~TabFact~}                                                      & \textbf{~WikiTQ~}                                                        & \textbf{~TabFact~}                                                      \\
      \hline \hline
		\revise{Off-Prep}  & \revise{${55.32}$}       & \revise{$81.67$}          & \revise{$56.86$}         & \revise{$\underline{81.52}$}          & \revise{$53.02$}         & \revise{$75.40$}          & \revise{$58.01$}         & \revise{$82.31$}          \\ \cline{1-9} 
		ICL-Prep  & ${56.54}$       & $80.53$          & $55.71$         & $73.91$          & $50.05$         & $75.20$          & $57.00$         & $80.14$          \\ 
      \hline \hline
		Dater     & $48.32$         & $83.05$          & $52.81$         & $72.08$          & $43.53$         & $74.01$          & $58.78$         & $79.84$          \\ \cline{1-9} 
		Binder    & $56.81$         & $82.81$          & ${56.74}$       & ${79.17}$        & $50.51$         & $78.16$          & $55.43$         & ${81.72}$        \\ \cline{1-9} 
		\revise{AutoTQA*}  & \revise{${60.80}$}       & \revise{$84.40$}          & \revise{$58.40$}         & \revise{$80.60$}          & \revise{$58.80$}         & \revise{$82.40$}          & \revise{$62.40$}         & \revise{${83.00}$}          \\ \cline{1-9} 
		ReAcTable & ${64.13}$       & ${85.71}$        & $51.80$         & $72.80$          & ${58.01}$       & ${80.00}$        & ${60.15}$       & $81.67$          \\ \cline{1-9} 
		CoTable   & $\underline{64.53}$       & $\underline{86.22}$        & $\underline{59.94}$       & ${80.20}$        & $\underline{62.22}$       & $\underline{85.62}$        & $\underline{64.41}$       & $\underline{83.20}$        \\
      \hline \hline
      {\sys}     & $\textbf{66.09}$       & $\textbf{87.85}$        & $\textbf{64.75}$       & $\textbf{84.19}$        & $\textbf{63.72}$       & $\textbf{85.72}$        & $\textbf{68.72}$       & $\textbf{85.33}$          \\
      \hline
    \end{tabular}
  }
  \label{tbl:vs_other_prep_method}
\end{table*}

\revise{

\noindent
\textbf{Exp-\expnum\label{exp:compare_dataprep_baseline}: Comparison of \sys with data prep baselines.}
We compare \sys against two representative baselines: a traditional offline data prep method Off-Prep and an LLM-based in-context learning data prep method ICL-Prep.

As shown in Table~\ref{tbl:vs_other_prep_method}, \sys consistently outperforms both baselines on WikiTQ and TabFact. Specifically, \sys improves over Off-Prep by \textbf{10.02} and \textbf{5.55} on average, respectively.  
The performance gap can be attributed to the fact that Off-Prep lacks question guidance, making it ineffective in addressing question-specific issues such as missing semantics and irrelevant columns. Moreover, predefined normalization routines struggle with table heterogeneity, 
whereas \sys can generate specialized code for such cases via LLMs.
Similarly, \sys outperforms ICL-Prep by \textbf{11.00} and \textbf{8.33} on average on the two datasets.  
This highlights that a multi-agent framework is essential for producing comprehensive data prep requirements and precise, executable programs.



\vspace{1mm}
\noindent
\textbf{Exp-\expnum\label{exp:compare_SOTA_TQA_with_prep}: Comparison of \sys with previous SOTA TQA methods with data prep.} 
We compare \sys with TQA methods that integrate data prep tasks in their question-answering process. The results, shown in Table~\ref{tbl:vs_other_prep_method}, highlight that \sys sets a new SOTA performance on both the WikiTQ and TabFact datasets across LLM backbones. Although CoTable achieves the best overall performance among existing SOTA methods, \sys outperforms it with an impressive average improvement of {\textbf{3.05}} on WikiTQ and {\textbf{1.96}} on TabFact. 
These improvements are largely attributed to our multi-agent framework, which effectively addresses the question-aware data prep challenges. Furthermore, when compared to AutoTQA, \sys shows significant gains of \textbf{5.81} on WikiTQ and \textbf{2.77} on TabFact. This is because AutoTQA lacks comprehensive data prep, leading to execution errors or incorrect answers.

The results demonstrate that data prep is inherently complex and cannot be solved with a one-size-fits-all solution. Instead, a more effective strategy involves specialized LLM-based agents for each type of data prep task, coordinated by a centralized planning agent. Moreover, \sys covers a broader range of data prep operations, filling gaps (\eg \term{Derive} and \term{Normalize}) that previous solutions have not fully addressed.

}





\revise{
Moreover, we also evaluate \sys on tables with various sizes. We find that, by employing multiple agents and program-based operations, \sys maintains stable performance as table size grows, ensuring that each data prep task is handled effectively without overwhelming a single model. More details on the experimental results and analysis can be found in our technical report~\cite{technical_report}.

As DeepSeek achieves the best accuracy at lower cost, we select it as our default backbone LLM for subsequent experiments.

}

\revise{

\subsection{Evaluation on Efficiency}
\label{subsec:efficiency_evaluation}


\begin{figure}[!t]
    \centering 
    \includegraphics[width=0.9\columnwidth]{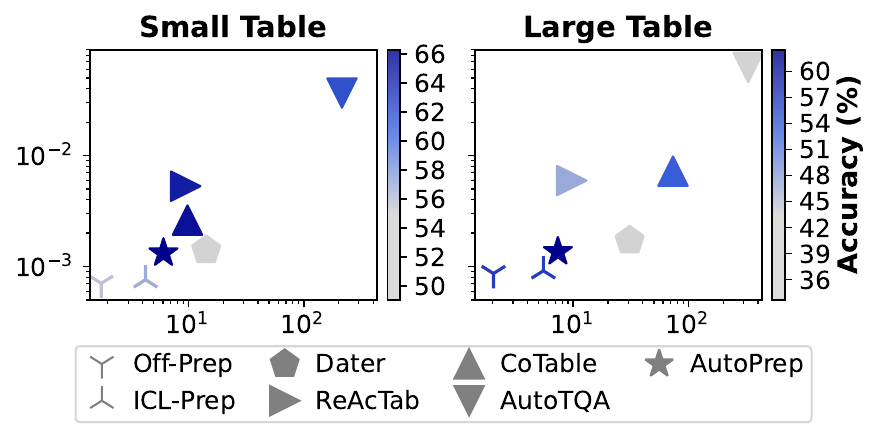}
    \vspace{-1em}
    \caption{\revise{Efficiency evaluation. 
    The $x$-axis and $y$-axis represent the time cost (in seconds) and monetary cost (in dollars) for processing a single instance, respectively. The color intensity of each scatter point reflects the overall accuracy.
    }
    }
    \label{fig:time_money_acc_scatter_map}
    \vspace{-1em}
\end{figure}

\noindent
\textbf{Exp-\expnum\label{exp:efficiency_evaluation}: How efficient is \sys compared with other methods?} 
To evaluate the efficiency of \sys, we compare it with other TQA methods in terms of time cost, monetary cost, and accuracy. The results for both small and large tables are shown in Figure~\ref{fig:time_money_acc_scatter_map}.
%
%
As shown, \sys achieves the best overall performance with insignificant time and monetary costs. Specifically, its cost is lower than all methods except Off-Prep and ICL-Prep. However, \sys significantly outperforms these two baselines in accuracy, achieving improvements of at least \textbf{9.36} and \textbf{5.77} points on small and large tables, respectively. Furthermore, as table size increases, \sys maintains a much more stable time and cost overhead compared to other state-of-the-art TQA methods.


This efficiency stems from the program-based data prep mechanism of \sys, which avoids high latency and API expenses associated with directly processing tables via LLMs. For instance, in the column derivation task on Table $T_1$ (Figure~\ref{fig:issue_instances_incomplete}), CoTable invokes LLMs to generate a full list of country codes. In contrast, \sys only generates a physical operator $\mathtt{extract}$ and executes it to produce the column, which results in significantly lower time and monetary costs, particularly on large tables.


%
%

}
\subsection{In-Depth Exploration of System Capabilities }
\label{subsec:evaluation_on_generalization}

\stitle{Exp-\expnum\label{exp:generalization_exp_on_table_bench}: Evaluating generalization on unseen datasets.} 
\revise{As discussed in Section~\ref{subsec:exp-setup}}, we evaluate the generalization capabilities of \sys on the TabBench datasets.
Specifically, we directly use the designed prompting strategies on the WikiTQ dataset, and examine whether these strategies can be generalized to TabBench.

\begin{figure}[!t]
    \centering 
    \includegraphics[width=0.9\columnwidth]{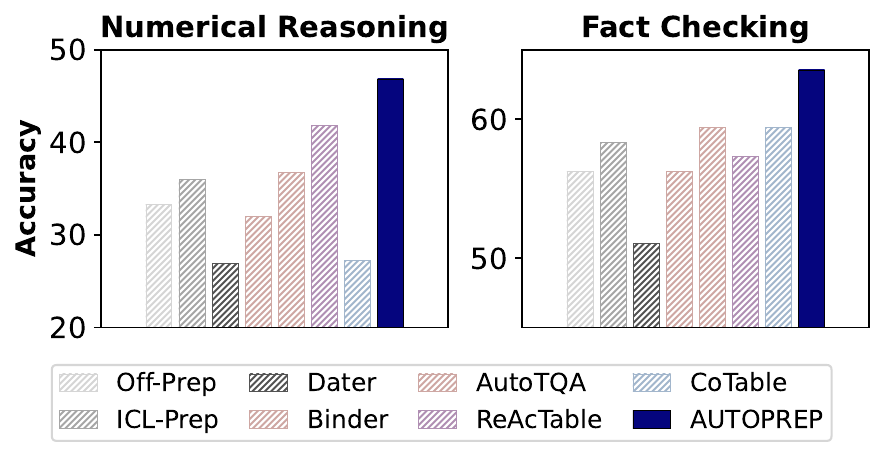}
    \vspace{-0.5em}
    \caption{\revise{Evaluating generalization on the TabBench dataset.}
    }
    \label{fig:tabbench_indepth_acc}
    \vspace{-1em}
\end{figure}

As shown in Figure~\ref{fig:tabbench_indepth_acc}, \sys achieves the highest overall accuracy among all methods. Specifically, compared with the second best method ReAcTable, \sys improves by
{\textbf{5.28}}, indicating the strong generalization capabilities of our method. 
The main reason is that \sys utilizes tool-augmented method for physical plan generation and execution, generalizing well on unseen datasets.

%

\revise{

\begin{table}[t]
  \centering
  \caption{\bf{
      \revise{
        Evaluating Extensibility for New Operation Types.
      }
    }}
  \vspace{-0.5em}
  \resizebox{0.7\linewidth}{!}{
    \begin{tabular}{|c||c|c|c|}
      \hline
      \multirow{2}{*}{\textbf{Method}}  & \multirow{2}{*}{\textbf{TransTQ}}  & \multicolumn{2}{c|}{\textbf{WikiTQ}} \\ \cline{3-4}
               &  & \textbf{~origin~} & \textbf{~after~}  \\
      \hline \hline
      ICL-Prep  & 58.92             & 56.54             & 55.00             \\ \cline{1-4} 
      ReAcTable  & 51.04             & \underline{64.13} & \underline{63.03} \\ \cline{1-4} 
      AutoTQA       & \underline{59.75} & 60.80             & 61.20             \\
      \hline \hline
      \sys        & \textbf{68.88}    & \textbf{66.09}    & \textbf{66.28}    \\ \hline
    \end{tabular}
  }
  \label{tbl:new_operator_new_operation}
  \vspace{-1em}
\end{table}
\noindent
\stitle{Exp-\expnum\label{exp:scalability_exp_new_op_type}: Evaluating extensibility on datasets requiring new logical operation types.} 
To further evaluate the extensibility of \sys, we evaluate it on a more complex dataset involving logical operations not originally supported. Since no existing TQA dataset presents sufficiently complex data quality issues, we construct a new dataset TransTQ with the assistance of LLMs, which is available on our GitHub repository. Specifically, we prompt an LLM to identify tables from WikiTQ where inverse transformation operations (\eg pivot and stack~\cite{li2023auto}) can be applied. These operations are then executed to generate non-relational table formats, introducing new challenges for TQA on {table transformation}.


As shown in Table~\ref{tbl:new_operator_new_operation}, without explicit guidance for handling table transformations, all baseline TQA methods exhibit poor accuracy. In contrast, after incorporating a new logical operation \texttt{Transform}, \sys achieves the best performance, outperforming the second-best method by \textbf{9.31} points. Moreover, we also validate that adding this new operation does not degrade the performance of \sys on the original WikiTQ dataset, confirming that its multi-agent architecture is well extensible to support the seamless integration of new data prep functionalities.



}

\subsection{Ablation Studies}
\label{subsec:ablation_study}

\noindent
\textbf{Exp-\expnum\label{exp:ablate_evaluate_on_planner}: Evaluation on the Planner agent.} 
We compare two Planner variants with different high-level operation suggestion methods, namely Direct Prompting and our proposed Chain-of-Clauses method, and report the results in Table~\ref{subtbl:planner_variants}.

We observe that Chain-of-Clauses outperforms Direct Prompting by \textbf{7.92}, \textbf{4.50}, \textbf{5.27} in accuracy on WikiTQ, TabFact and TabBench respectively.
This indicates the superiority of our proposed method in generating more accurate logical operations.
Moreover, we find that the performance improvement on WikiTQ is more significant than that on other datasets. 
This is because the WikiTQ dataset has relatively large
tables and complex questions, which could make the logical
operation suggestion problem more challenging to be solved using
Direct Prompting.

%


\stitle{Exp-\expnum\label{exp:ablate_evalaute_on_programmer_agents}: Evaluation on the Programmer Agents.} 
We evaluate the low-level operation generation methods in the 
\textsc{Programmer} agents and keep other settings of \sys as default.

As illustrated in Table~\ref{subtbl:programmer_varients}, our proposed tool-augmented method achieves better performance compared with the code generation method.
Considering the logical operations input to the Programmer agents are the same, we can conclude that selecting a function from a function pool and then completing its arguments can generate more accurate and high-quality programs to implement the high-level operations. Moreover, when using demonstrations constructed from same table-question pairs for the prompt of programmer agents, Tool-augmented method can save 18.07\% input tokens for low-level operation generation. 
We also record the error ratio of these two methods, as shown in Table~\ref{subtbl:programmer_varients}. The probability of bugs in our tool-augmented method is greatly reduced (\eg from 4.51\% to 1.50\%), demonstrating its effectiveness.

%

\begin{table}[t]
    \centering
    \caption{\bf{Experimental Results of Ablation Studies.}}
    \vspace{-1em}
    \label{tbl:ablation_study_main_table}
    
    \begin{subtable}[t]{\columnwidth}
        \centering
        \caption{\bf{Evaluation on the \textsc{Planner} agent.}}
        \label{subtbl:planner_variants}
        \vspace{-0.15em}
        \resizebox{0.93\linewidth}{!}{
        \begin{tabular}{|c||c|c|c|}
            \hline
            \textbf{Method} & ~\textbf{WikiTQ}~ & ~\textbf{TabFact}~ & ~\textbf{TabBench}~ \\ \hline \hline
            Direct Prompting & $58.17$ & $83.35$ & $44.83$ \\ \hline
            Chain-of-Clauses & $\textbf{66.09}$ & $\textbf{87.85}$ & $\textbf{50.10}$ \\ 
            \hline
        \end{tabular}
        }
    \end{subtable}
    
    \vspace{0.30em}
    
    \begin{subtable}[t]{\columnwidth}
        \centering
        \caption{\bf{Evaluation on the \textsc{Programmer} Agents.
        }}
        \label{subtbl:programmer_varients}
        \vspace{-0.15em}
        \resizebox{0.93\linewidth}{!}{
        \begin{tabular}{|c|c||c|c|c|}
            \hline
            \textbf{Method} & \textbf{Metric} & ~\textbf{WikiTQ}~ & ~\textbf{TabFact}~ & ~\textbf{TabBench}~ \\ \hline \hline
            \multirow{2}{*}{\makecell[c]{{Code}\\{Generation}}} & Acc $\uparrow$ & $62.82$ & $81.97$ & $47.26$ \\ \cline{2-5}
             & Err $\downarrow$ & $4.51\%$ & $4.50\%$ & $4.73\%$ \\ \hline
            \multirow{2}{*}{\makecell[c]{{Tool}\\{Augmented}}} & Acc $\uparrow$ & $\textbf{66.09}$ & $\textbf{87.85}$ & $\textbf{50.10}$ \\ \cline{2-5}
             & Err $\downarrow$ & $\textbf{1.50\%}$ & $\textbf{0.05\%}$ & $\textbf{1.01\%}$ \\
            \hline
        \end{tabular}
        }
    \end{subtable}
    
    \vspace{0.30em}
    
    \begin{subtable}[t]{\columnwidth}
        \centering
        \caption{\bf{Contribution of Each \textsc{Programmer} Agent in \sys.}}
        \label{subtbl:ablation_exp}
        \vspace{-0.15em}
        \resizebox{0.93\linewidth}{!}{
        \begin{tabular}{|c||c|c|c|}
            \hline
            \textbf{Method} & ~\textbf{WikiTQ}~ & ~\textbf{TabFact}~ & { ~\textbf{TabBench}~} \\ \hline \hline
            \sys & $\textbf{66.09}$ & $\textbf{87.85}$ & { $\textbf{50.10}$} \\ \hline \hline
            - \textsc{Filter} & $62.78$ {\scriptsize \textcolor{dropred}{($-3.31$)}} & $84.98$ {\scriptsize \textcolor{dropred}{($-2.87$)}} & {$47.67$} {\scriptsize \textcolor{dropred}{($-2.43$)}} \\ \hline
            - \textsc{Derive} & $61.79$ {\scriptsize \textcolor{dropred}{${(-4.30)}$}} & $85.67$ {\scriptsize \textcolor{dropred}{($-2.18$)}} & {$45.44$} {\scriptsize \textcolor{dropred}{($-4.66$)}} \\ \hline
            - \textsc{Normalize} & $62.02$ {\scriptsize \textcolor{dropred}{($-4.07$)}} & $83.79$ {\scriptsize \textcolor{dropred}{${(-4.06)}$}} & {$41.58$} {\scriptsize \textcolor{dropred}{${(-8.52)}$}} \\ \hline
        \end{tabular}
        }
    \end{subtable}
    
    \vspace{-1em}
\end{table}

\stitle{Exp-\expnum\label{exp:ablate_contribute_of_each_agent}: Contribution of each agent in \sys.} 
We ablate each \textsc{Programmer} agent including \textsc{Filter}, \textsc{Derive} and \textsc{Normalize} and compare the performance with \sys.

As shown in Table~\ref{subtbl:ablation_exp}, for WikiTQ, without column derivation, the accuracy drops the most ({4.30}). For TabFact and TabBench, the Normalize matters the most with an accuracy drop by {4.06 and 8.52}. This is because that WikiTQ has more instances requiring string extraction or calculation to generate new columns for answering the question, while normalization is a primary issue in TabFact and TabBench. Moreover, for all datasets, each agent plays an essential role in data preparation for TQA, which brings accuracy improvement by at least \textbf{2.43}, \textbf{2.18} and \textbf{4.06}.


\section{Related Work}
\label{sec:related_work}

\stitle{Tabular Question Answering. }
Most of the SOTA solutions for TQA rely on LLMs~\cite{DBLP:conf/nips/BrownMRSKDNSSAA20, DBLP:journals/corr/abs-2401-14196}, as TQA requires NL understanding and reasoning over tables. 
There are two types of methods for TQA named \emph{Direct Prompting}~\cite{chen2022large, wei2022chain} and \emph{Code Generation}~\cite{rajkumar2022evaluating}. 
%
Previous TQA methods, like Dater~\cite{ye2023large}, Binder~\cite{cheng2023binding}, CoTable~\cite{wang2024chain} and ReAcTable~\cite{zhang2024reactable} also consider data prep in their question answering process.
%
Specifically, Dater~\cite{ye2023large} prompts LLMs to select relevant columns related to answering the question, targeted at solving filtering tasks. 
Binder~\cite{cheng2023binding} proposes to integrate SQL with an LLM-based API to incorporate external knowledge, which may partly address derivation tasks. 
Similarly, CoTable~\cite{wang2024chain} addresses the filtering tasks and derivation tasks by designing operators which are implemented by LLM completion.
ReAcTable~\cite{zhang2024reactable} uses few-shot demonstrations to instruct LLMs to generate python code or SQL to address the derivation and filtering tasks.

However, previous TQA methods do not address the column normalization tasks, which account for the most significant error types, as indicated in Figure~\ref{fig:error_instance_record}. 
Second, the methods utilize a single LLM agent for both data prep and answer reasoning. Given that data prep is a complex challenge, these one-size-fits-all solutions may not achieve satisfactory performance. 

\stitle{Traditional Data Preparation.}
Data prep techniques are widely used across various tasks~\cite{DBLP:conf/sigmod/Chai0FL23, DBLP:conf/icde/FanHFC00024, DBLP:conf/sigmod/TuF0WC0F022}. For training machine learning models for data analytics, 
%
Auto-Weka~\cite{thornton2013auto} leverages Bayesian optimization to identify data prep operations.
Auto-Sklearn~\cite{ feurer2020auto, hutter2019automated} and TensorOBOE~\cite{ DBLP:conf/kdd/YangFWU20} apply meta-learning to discover promising operations. Alpine Meadow~\cite{ DBLP:conf/sigmod/ShangZBKECBUK19} introduces an exploration-exploitation strategy, while TPOT~\cite{olson2016tpot} uses a tree-based representation of data prep and genetic programming optimization techniques. Several studies~\cite{ DBLP:conf/www/Berti-equille19, DBLP:conf/sigmod/ElMS20, DBLP:conf/kdd/HeffetzVKR20, zhang2025reward} explore reinforcement learning techniques. 
HAIPipe~\cite{ DBLP:journals/pacmmod/Chen0FYCL023} integrates both human-orchestrated and automatically generated data prep operations. 

\revise{We propose \sys to generate \emph{question-aware} data prep operations for TQA tasks. 
Traditional data prep methods typically operate at an {\it offline stage}, independent of any specific downstream question or query. In contrast, question-aware data prep studied in \sys {tailors the tables to the specific needs of the question} during the {\it online stage}, directly addressing the challenge of aligning the table’s structure with the NL question's semantics. }

\stitle{LLMs-based Multi-Agent Framework.}  A multi-agent LLM framework refers to a well-designed hierarchical structure consisting of multiple LLM-based agents and scheduling algorithms~\cite{han2024llm}. Compared with single-agent methods based on prompting techniques, such structures are better suited for handling complex tasks like software development, issue resolution, and code generation~\cite{hong2023metagpt, qian2023communicative, chen2024coder, tao2024magis,ishibashi2024self, islam2024mapcoder}. 
\revise{The structure of multi-agent frameworks can be categorized into equi-level~\cite{terekhov2023second}, hierarchical~\cite{ahilan2019feudal, harris2023stackelberg}, and nested~\cite{chan2023chateval} structures.}
AutoTQA~\cite{DBLP:journals/pvldb/ZhuCXLSZSTL24} proposes a multi-agent framework with hierarchical structure for supporting Tabular Question Answering. 

\revise{
We adopt the hierarchical structure because our design introduces a Planner agent that decomposes the overall data preparation task into three distinct sub-tasks, each handled by specialized agents. This top-down coordination naturally aligns with the principles of the hierarchical multi-agent framework.
Moreover, while AutoTQA focuses on improving table analysis, this paper focuses on the performance bottleneck caused by data prep issues in TQA. Our proposed framework, \sys, addresses question-aware data prep for TQA and can be integrated as a plugin into current TQA approaches to further improve the overall performance.
}

\section{Conclusion and Future work}\label{sec:conclusion_and_future_work}

In this paper, we have introduced \sys, an LLM-based multi-agent framework to support data prep for TQA tasks. \sys consists of three stages: (1) the Planning stage, which suggests logical data prep operations, (2) the Programming stage, which generates physical implementations for each logical operation, and (3) the Executing stage, which executes the Python code and reports error messages. We propose a Chain-of-Clauses method to generate high-quality logical plans and a Tool-augmented method for effective physical plan generation. Extensive experiments on real datasets demonstrate the superiority of \sys.


\revise{
For future work, we identify three promising directions. First, we aim to extend the system's data prep capabilities by integrating a broader set of operations to handle more complex issues such as missing values, duplicates, etc. Second, \sys currently focuses on single-table question answering, and extending its capabilities to support multi-table TQA is essential for tackling more realistic and complex scenarios.
The third direction is to enable question-aware data preparation over enterprise datasets, which are more complex than existing TQA benchmarks. 
%
}
\section*{Acknowledgment}\label{sec:ack}


This paper was partly supported by the NSF of China (62436010, 62441230, 62525202 and 62232009), National Key R\&D Program of China (2023YFB4503600), NSF DBI-2327954, Guangdong Provincial Project 2023CX10X008 and Amazon Research Awards.

\newpage
\bibliographystyle{ACM-Reference-Format}
\bibliography{citations/ref} 
\balance

\end{document}